\documentclass[12pt]{article}
\pdfoutput=1 
\usepackage{amsmath}
\usepackage{amssymb}
\usepackage{amsfonts}
\usepackage{epsfig}
\usepackage{mathrsfs}
\usepackage{amsmath,amsthm,amssymb,amscd}
\usepackage{MnSymbol}
\usepackage{rotating}
\usepackage{multirow}
\usepackage{float}
\usepackage[english]{babel}
\usepackage{floatpag}
\usepackage{enumerate}
\usepackage{threeparttable}
\usepackage{algorithm}
\usepackage{algorithmic}
\usepackage{float}
\usepackage{graphicx}
\usepackage{booktabs}
\usepackage{xcolor}

\newcommand\appendix@section[1]{\refstepcounter{section}\orig@section*{#1}\addcontentsline{toc}{section}{#1}}
\let\orig@section\section\g@addto@macro\appendix{\let\section\appendix@section}
\makeatother \makeatletter
\renewcommand\footnoterule{\kern-3\p@ \hrule width 1\columnwidth \kern 2.6\p@}
\newskip\@footindent
\renewcommand\@footindent{0pt}
\@addtoreset{footnote}{page}
\long\def\@makefntext#1{\@setpar{\@@par\@tempdima \hsize
\advance\@tempdima-\@footindent \parshape \@ne \@footindent
\@tempdima}\par \noindent \hbox to
\z@{\hss\@thefnmark��\hspace{0.2em}}#1}

\def\@makefnmark{\hbox{\textsuperscript{\@thefnmark}}}
\makeatother

\newtheorem{thm}{Theorem}

\title{Efficient semi-supervised inference for logistic regression under case-control studies}
\author{ \vspace{2mm} Zhuojun Quan\\
	Department of Statistics and Data Science\\
	School of Management\\
	Fudan University\\
	\vspace{10mm}
	Shanghai, P.R.China\\
	\and
	\vspace{2mm} Yuanyuan Lin\\
	Department of Statistics\\
	The Chinese University of Hong Kong\\
	\vspace{10mm}
	Shatin, New Territories, Hong Kong\\
	\and
	\vspace{2mm} Kani Chen\\
	Department of Mathematics\\
	Hong Kong University of Science and Technology\\
	\vspace{10mm}
	Clear Water Bay, Kowloon, Hong Kong\\
	\and
	\vspace{2mm} Wen Yu\\
	Department of Statistics and Data Science\\
	School of Management\\
	Fudan University\\
	\vspace{10mm}
	Shanghai, P.R.China
}
\date{ }
\begin{document}
\maketitle
\newpage
\begin{abstract}
Semi-supervised learning has received increasingly attention in statistics and machine learning. In semi-supervised learning settings, a labeled data set with both outcomes and covariates and an unlabeled data set with covariates only are collected. We consider an inference problem in semi-supervised settings where the outcome in the labeled data is binary and the labeled data is collected by case-control sampling. Case-control sampling is an effective sampling scheme for alleviating imbalance structure in binary data. Under the logistic model assumption, case-control data can still provide consistent estimator for the slope parameter of the regression model. However, the intercept parameter is not identifiable. Consequently, the marginal case proportion cannot be estimated from case-control data. We find out that with the availability of the unlabeled data, the intercept parameter can be identified in semi-supervised learning setting. We construct the likelihood function of the observed labeled and unlabeled data and obtain the maximum likelihood estimator via an iterative algorithm. The proposed estimator is shown to be consistent, asymptotically normal, and semiparametrically efficient. Extensive simulation studies are conducted to show the finite sample performance of the proposed method. The results imply that the unlabeled data not only helps to identify the intercept but also improves the estimation efficiency of the slope parameter. Meanwhile, the marginal case proportion can be estimated accurately by the proposed method.
\end{abstract}

\vfill \hrule \vskip 6pt \noindent {\em MSC:}  \ \\
\noindent {\em Some key words}: Biased sampling; Case-control study; Classification; Maximum likelihood estimator; Semiparametric efficiency; Semi-supervised inference.

\newpage

\section{Introduction}
 
Logistic regression (Cox, 1958) is probably the most fundamental statistical  model to fit binary or multi-category outcome. Let $Y$ be the binary response variable taking values $0,1$ and let $X$ be an observable $p$-dimensional vector of covariates. A binary logistic regression assumes that
\begin{equation}\label{M1}
\mathsf{P}(Y=1 \mid X=x)=1-\mathsf{P}(Y=0 \mid X=x)=\frac{\exp(\alpha+\beta^{\top} x)}{1+\exp(\alpha+\beta^{\top} x)},
\end{equation}
where $\beta \in{\mathbb R}^{p}$ is the slope parameter and $\alpha \in{\mathbb R}$ is the intercept. The logistic regression is often regarded as a special case of the generalized linear models  with logit link function. 
Model (\ref{M1}) is particularly popular in the analysis of medical and epidemiological data for the study of the effect of certain exposure to possible disease or hazards. Usually if disease is present $Y=1$ (cases), else $Y=0$ (controls). Prospective studies follow a sample of subjects or individuals and the record corresponding $Y$. For logistic regression model under prospective studies, the samples are random samples from the underlying population. 
 
Nonetheless, many diseases are rare in medical studies,  and large prospective
study might produce very few disease cases and thus very little information about the disease of interest. Due to the rareness of some diseases, it is sometimes impractical to study them by following up of an initially healthy population, for example  the study of smoking and lung cancer.  Retrospective study is an effective tool for the study of the effect of certain exposure to rare disease incidence, which enjoys important advantages over prospective study. Case-control study is a special retrospective study, by taking separate random samples from the case population and the control population, when the case population and the control population can be clearly separated  (Chen and Lo, 1999).   In general, case-control sample is a special case of the so-called response biased sampling, where the sampling schemes rely on the value of outcome; see Chen(2001), Manski and Lerman (1977),  Cosslett (1981), Scott and Wild (1986,1997), Fithian and Hastie (2014), and Yao {\it et al. }(2017). For logistic regression under case-control study, a remarkable finding is that, the prospective estimating equation  is valid for a consistent estimate of the slope parameter $\beta,$ except the intercept term $\alpha$ (Prentice and Pyke, 1979).  In classical logistic regression under case-control study, the intercept term and the density function of $X$ are not
identifiable/estimable, as 
the score function of the intercept  $\alpha$ lies in the linear space spanned by the score function of the density function of $X$. 
The identifiability/estimability  of  logistic regression under (multiple) case-control studies has been carefully studied by Tang {\it et al.} (2021). 
    
In recent years, semi-supervised learning has gained significant attentions in  statistics and machine learning.
Under semi-supervised setting, there are typically two sources of data: (i) A relatively small  labeled data set collected according to certain sampling scheme, denoted by $\mathcal{L}$, containing observations of the outcome variable $Y \in \mathbb{R}$ and the covariates $X\in\mathbb{R}^p$. (ii) A relatively large unlabeled data set, denoted by $\mathcal{U}$, containing only observations of covariate $X$. Semi-supervised setting is 
 rather common  in practical applications, as the outcome variable is often more costly or difficult to collect than the predictors (Chapelle {\it et al.,} 2009). For example, in analyzing the electronic health records (EHR) data, labeling of the outcome variable is often very laborious  whereas the covariates are relatively easy to obtain (Gronsbell and Cai, 2018; Chakrabortty and Cai, 2018). Earlier works on semi-supervised learning focus on classification problems (Nigam {\it et al.,} 2000; Ando and Zhang, 2005, 2007;  Belkin {\it et al.,} 2006; Wang and Shen, 2007; Wang {\it et al.,} 2008; etc). Semi-supervised learning in regression framework were reported by Belkin {\it et al.} (2006), Wasserman and Lafferty (2007), Johnson and Zhang (2008), among many others.  A comprehensive review on related works can be found in
Zhu and Goldberg (2009).

How to integrate the labeled and unlabeled data for a joint analysis, that can produce more accurate and efficient parameter estimation, has become one of central concerns in semi-supervised learning. This is known as semi-supervised inference in statistics. Let $\mathbb{P}$ be the joint distribution of $(Y, X^{\top})^{\top}$ and let $\mathbb{P}_{X}$ be the marginal distribution of $X$. And let $\theta^{\star}\equiv \theta^{\star}(\mathbb{P})$ be a target parameter of interest. Generally speaking, whether the unlabeled data can be used to improve the estimation efficiency of $\theta^{\star}$ rests on whether $\theta^{\star}$ is explicitly or implicitly related to  $\mathbb{P}_{X}$, as acquiring a large unlabeled data set is approximately equivalent to knowing $\mathbb{P}_{X}$ (Zhang and Oles, 2000). Recently,  important works on semi-supervised inference without any parametric  model assumptions relating $X$ and $Y$ were reported.  Azriel {\it et al.} (2016) considered the construction of the best linear predictor in semi-supervised framework. Semi-supervised inference of the population mean of a continuous response variable  was studied by Zhang {\it et al.} (2019). Chakrabortty and Cai (2018) proposed a class of efficient and adaptive estimators to improve estimation efficiency in semi-supervised settings. Efficient and robust semi-supervised estimation of the averaged treatment effects was studied by Cheng {\it et al.} (2018). Efficient evaluation of model predictive performance with semi-supervised methods was considered by Gronsbell and Cai (2018). For high dimensional linear models, novel 
semi-supervised calibrated estimator for the explained variance  was developed by Cai and Guo (2020).

In this paper,  we study an efficient semi-supervised inference  procedure for the logistic regression, where the labeled data is collected by case-control sampling.  With the availability of the unlabeled data,   the intercept parameter becomes identifiable under mild conditions, contrary to classical case-control logistic regression; more importantly, the resulting estimate of the slope parameters are proved to be semiparametric efficient. By maximizing the nonparametric likelihood function of the combined data, the resulting estimators for the intercept and slope parameters are shown to be consistent, asymptotically normal and asymptotically efficient. An iterative algorithm is proposed to compute the maximum likelihood estimators numerically.   
      
The rest of the paper is organized as follows. In Section 2, we introduce some necessary notation and the structure of the observed data. In Section 3, we first discuss the identifiability issue of the model parameters. Then the likelihood function is derived to estimate the parameters and the iterative algorithm is given out. We also show the large sample properties of the resulting estimators. The numerical results are presented in Section 4, including extensive simulation studies and the analysis of the Pima Indians diabetes data. Section 5 concludes. All the technical details are summarized in the Appendix.

\section{Notation and data structure}
Using the notation defined in Section 1, we call the class with $Y=1$ case population and the class with $Y=0$ control population. Define the function 
\begin{eqnarray*}
\phi(x,\theta)=\frac{\exp(\alpha+\beta^\top x)}{1+\exp(\alpha+\beta^\top x)},
\end{eqnarray*}
where $\theta=(\alpha,\beta^\top)^\top$. Then the binary logistic regression model (\ref{M1}) can be written as $\mathsf{P}(Y=1|X=x)=\phi(x,\theta)$.

The structure of the observed data is as follows. We have a random sample of size $n_1$ from the case population and a random sample of size $n_0$ from the control population. The two samples compose the labeled data of size $n=n_1+n_0$. Besides the labeled data, we collect a random sample from $\mathbb P_X$, i.e., the marginal distribution of $X$, with size $N-n$, which is treated as the unlabeled data. We denote the labeled data by ${\cal D}_{\cal L}=\{(y_i,x_i), i=1,\ldots,n\}$ and the unlabeled data by ${\cal D}_{\cal U}=\{x_i, i=n+1,\ldots,N\}$. 

Assume that $\mathbb P_X$ has a density. Denote the density function and the cumulative distribution function of $X$ by $f(x)$ and $F(x)$, respectively. Let $f_1(x)$ be the conditional density function of $X$ given $Y=1$ and $f_0(x)$ be the one given $Y=0$. Under the logistic model (\ref{M1}), using the Bayes formula, it is not difficult to show that 
\begin{eqnarray}\label{2.2}
f_1(x)=\frac{\phi(x,\theta)f(x)}{c(\theta,F)}, \ f_0(x)=\frac{(1-\phi(x,\theta))f(x)}{1-c(\theta,F)},
\end{eqnarray}
where $c(\theta,F)=\int_{\cal X}\phi(x,\theta)dF(x)$ with ${\cal X}$ being the sample space of the covariates $X$.
 
\section{Main results}

\subsection{Identifiability issue} 

As we have mentioned, the case-control studies are quite common for analyzing binary response data to alleviate obvious imbalance between cases and controls. Essentially, the case-control sampling is a biased sampling scheme. However, Farewell (1979) argued that if the logistic model (\ref{M1}) holds for $Y$ and $X$, then for the case-control sampling data, the observed conditional probability of $Y=1$ given $X$ still satisfies the form of (\ref{M1}) with the same $\beta$ but a different $\alpha$. That is, the slope parameter $\beta$ in (\ref{M1}) is still identifiable for a single case-control sample. As we have mentioned in Section 1, Prentice and Pyke (1979) showed that by fitting a usual prospective logistic model for the case-control data, one can still get a consistent estimator for $\beta$ and the corresponding inference procedure is also valid. Contrary to $\beta$, the intercept parameter $\alpha$ is not identifiable under a single case-control sample. Although $\beta$ is more important to understand the relationship between $Y$ and $X$, $\alpha$ is also useful for accurate classification and prediction. To identify $\alpha$, some extra information outside the case-control sample is needed. 

For the semi-supervised setting, we shall always assume the sizes of cases and controls, i.e., $n_1$ and $n_0$, and the size of the unlabeled data, i.e., $N-n$, tend to infinity. Meanwhile, we assume that the support of $X$ is not limited to any hyperplane whose dimension is smaller than $p$. Under these assumptions, $f(x)$ is identifiable as $N-n$ goes to infinity. Knowing from the classical result of case-control studies, $\beta$ is identifiable as long as $n_1$ and $n_0$ tend to infinity. Let $x_1$ and $x_2$ be two different values in the support of $X$ such that  $\beta^\top x_1\not=\beta^\top x_2$. From (\ref{2.2}), we have that $f_0(x_1)/f(x_1)=(1-\phi(x_1,\theta))/(1-c(\theta,F))$ and $f_0(x_2)/f(x_2)=(1-\phi(x_2,\theta))/(1-c(\theta,F))$. From these two quantities, we can solve out $\alpha$ as long as $\beta^\top x_1\not=\beta^\top x_2$. Thus, $\alpha$ is identifiable when $\beta\not=0$. If $\beta=0$, it is clear that different $\alpha$ values give the same case and control densities which are the same as $f(x)$. That means $\alpha$ is not identifiable if $\beta=0$.

\subsection{Maximum likelihood estimation} 

To estimate the unknown parameters, we use the maximum likelihood estimation approach. The likelihood function of the observed data is given by
\begin{equation*}
\prod_{i=1}^n f_1(x_i)^{y_i}f_0(x_i)^{1-y_i}\prod_{i=n+1}^Nf(x_i).
\end{equation*}
Using the relationship in (\ref{2.2}) and taking the logarithm of the likelihood, we obtain the following log-likelihood function of $\theta$ and $f$
\begin{eqnarray*}
l(\theta,f)&=&\sum_{i=1}^n\left[y_i\log\phi(x_i,\theta)+(1-y_i)\log(1-\phi(x_i,\theta))\right]\\
& &-\left[n_1\log c(\theta,F)+n_0\log(1-c(\theta,F))\right]+\sum_{i=1}^N\log f(x_i).
\end{eqnarray*}
To deal with the non-parametric component $f$, we adopt the discretization approach in Zeng and Lin (2007) and many others. Specifically, assume that there is no tie in the observed values of $X$. We restrict $F$ to be a step function with non-negative jumps only at the observed $x_i$'s. Let $p_i$ be the jump size of $F$ at $x_i$, $i=1,\ldots,N$. Accordingly, $p_i$'s satisfy that $p_i\geqslant0$ and $\sum_{i=1}^Np_i=1$. Meanwhile, by this restriction, $c(\theta,F)=\sum_{i=1}^N\phi(x_i,\theta)p_i$. To distinguish from its original definition, we use the notation $\tilde{c}(\theta,{\bf p})$ to represent this discretization version instead, where ${\bf p}=(p_1,\ldots,p_N)$.

After the discretization, the log-likelihood function becomes
\begin{eqnarray*}
l(\theta,{\bf p})&=&\sum_{i=1}^n\left[y_i\log\phi(x_i,\theta)+(1-y_i)\log(1-\phi(x_i,\theta))\right]\\
& &-\left[n_1\log\tilde{c}(\theta,{\bf p})+n_0\log(1-\tilde{c}(\theta,{\bf p}))\right]+\sum_{i=1}^N\log p_i,
\end{eqnarray*}
where $\tilde{c}(\theta,{\bf p})=\sum_{i=1}^N\phi(x_i,\theta)p_i$, as defined before. We maximize the log-likelihood $l(\theta,{\bf p})$ with respect to $\theta$ and $\bf{p}$ under the constraints $\sum_{i=1}^Np_i=1$ and $p_i\geqslant0, i=1,\ldots,N$. Denote the maximizer by $\hat{\theta}$ and $\hat{{\bf p}}=(\hat{p}_1,\ldots,\hat{p}_N)$. Define the estimator $\hat{F}(x)=\sum_{x_i\in\mathbf{r}(x)}\hat{p}_i$, where $\mathbf{r}(x)$ is the rectangular set decided by $x$. Then $(\hat{\theta}, \hat{F}(x))$ is treated as the maximum likelihood estimator (MLE) of $(\theta,F(x))$.

Based on the logistic model (\ref{M1}) and the MLEs of the model parameters, a natural estimator for the case proportion, i.e., $\mathsf{P}(Y=1)$, is given by $\hat{P}=\sum_{i=1}^N\phi(x_i;\hat{\theta})\hat{p}_i$.

\subsection{The algorithm}

The optimization problem is
\begin{equation}\label{3.2}
\max_{\theta,{\bf p}}l(\theta,{\bf p}) \ \ \ \mbox{subject to} \ \ \ \sum_{i=1}^Np_i=1, p_i\geqslant 0, i=1,\ldots,N.
\end{equation}
Motivated by the idea of coordinate descent, we develop an iterative algorithm to solve this constrained optimization problem (\ref{3.2}). 

Specifically, let ${\bf p}^{(0)}=(p_1^{(0)},\ldots,p_N^{(0)})$ be an initial value of $\bf p$. For the $j$-th ($j\geqslant0$) step of the iteration, given the value of ${\bf p}^{(j)}$, we update the value of $\theta$ by maximizing 
\begin{eqnarray}\label{3.3}
\sum_{i=1}^n\left[y_i\log\phi(x_i,\theta)+(1-y_i)\log(1-\phi(x_i,\theta))\right]
-\left[n_1\log\tilde{c}^{(j)}(\theta)+n_0\log(1-\tilde{c}^{(j)}(\theta))\right]
\end{eqnarray}
with respect to $\theta$, where $\tilde{c}^{(j)}(\theta)=\sum_{i=1}^N\phi(x_i,\theta)p_i^{(j)}$. Denote the maximizer by $\theta^{(j)}$. The optimization can be solved by the gradient descent algorithm.

Given the value of $\theta^{(j)}$ and ${\bf p}^{(j)}$, we update the value of $\bf p$ by maximizing $l(\theta^{(j)},{\bf p})$ with respect to $\bf p$, subject to the constraints $\sum_{i=1}^Np_i=1$ and $\sum_{i=1}^N\phi(x_i,\theta^{(j)})p_i=\tilde{c}^{(j)}(\theta^{(j)})$. This is a convex optimization problem. The unique solution is given by
\begin{equation*}
p_i=\left(\frac{n_1}{\tilde{c}^{(j)}(\theta^{(j)})}\phi(x_i,\theta^{(j)})+\frac{n_0}{1-\tilde{c}^{(j)}(\theta^{(j)})}\left(1-\phi(x_i,\theta^{(j)})\right)+N-n_1-n_0\right)^{-1},
\end{equation*}
$i=1,\ldots,N$. Thus, the value of $p^{(j)}$'s is updated by
\begin{equation}\label{3.4}
p_i^{(j+1)}=\left(\frac{n_1}{\tilde{c}^{(j)}(\theta^{(j)})}\phi(x_i,\theta^{(j)})+\frac{n_0}{1-\tilde{c}^{(j)}(\theta^{(j)})}\left(1-\phi(x_i,\theta^{(j)})\right)+N-n_1-n_0\right)^{-1},
\end{equation}
$i=1,\ldots,N$. We iteratively update the estimate values by maximizing (\ref{3.3}) and calculating (\ref{3.4}) until convergence. The resulting values are treated as the MLEs.

The iterative algorithm is summarized in Algorithm 1.
\begin{algorithm}
\caption{Iterated algorithm for computing the MLEs }  
\label{}  
\begin{algorithmic}  
\STATE {Set an initial value $\textbf{p}^{(0)}.$}   
\REPEAT 
\STATE Step 1. Given $\textbf{p}^{(j)}$, compute $\theta^{(j)}$ by maximizing ($\ref{3.3}$).
\STATE Step 2. Given $(\theta^{(j)}, \textbf{p}^{(j)})$, obtain $\textbf{p}^{(j+1)}$ according to (\ref{3.4}).
\UNTIL{$\|\theta^{(j+1)}-\theta^{(j)}\|\leqslant\epsilon$ and $\|\textbf{p}^{(j+1)}-\textbf{p}^{(j)}\|\leqslant\epsilon$, where $\|\cdot\|$ is the Euclidean norm and $\epsilon$ is a given constant.}	
\end{algorithmic} 
\end{algorithm}

At each iteration, the value of the objective function $l(\theta,{\bf p})$ increases. Since $l(\theta,{\bf p})$ is continuous in $\theta$ and $\bf p$ and is bound by $l(\hat{\theta},\hat{\bf p})$, the proposed iterative algorithm will at least converge to a local maximizer. To obtain the global maximizer with higher probability, one may try different initial values. 

\subsection{Large-sample properties}

We discuss the asymptotic properties of the proposed MLE $\hat{\theta}$ and $\hat{F}(x)$. Let $\theta_0=(\alpha_0,\beta_0^\top)^\top$ and $F_0(x)$ be the true value of $\theta$ and $F(x)$, respectively. Let $\mathcal{G}=\{g\in\mbox{BV}[\mathbb{R}^{d}]: |g|\leqslant1\}$, where $\mbox{BV}[D]$ is the set of functions on domain $D$ with bounded total variation. The following regularity conditions are imposed.

\medskip

\noindent{\it C1}. \ The density $f(x)$ is continuous with bounded support.

\noindent{\it C2}. \ If $\mathsf{P}(u^\top X=0)=1$ for some constant vector $u$, then $u=0$.

\noindent{\it C3}. \ The true parameter value $\theta_0\in{\mathcal B}$, where ${\mathcal B}$ is a compact set.

\medskip

The condition $\it C1$ implies that the covariates are bounded. The condition $\it C2$ means that the covariates vector does not lie in a degenerated linear space. The condition $\it C3$ assumes that $\theta_0$ is an interior point of a compact parameter space. They are all common regularity conditions.

The main results includes the consistency and the asymptotic normality of the MLEs, as given in the following two theorems.

\begin{thm}\label{thm1}
(Consistency) Suppose that conditions C1-C3 hold. If $n/N\to c_1$ for some constant $c_1\in(0,1)$ and $n_1/n_0\to c_2$ for some constant $c_2\in(0,\infty)$ as $N\to\infty$, then $|\hat{\theta}-\theta_0|\to0$ and $\sup_{x\in\mathbb{R}^p}|\hat{F}(x)-F_0(x)|\to0$ almost surely.
\end{thm}

\begin{thm}\label{thm2}
(Asymptotic normality) Suppose that conditions C1-C3 hold. If $n/N\to c_1$ for some constant $c_1\in(0,1)$ and $n_1/n_0\to c_2$ for some constant $c_2\in(0,\infty)$ as $N\to\infty$, then $\sqrt{N}(\hat{\theta}-\theta_0, \hat{F}-F_0)$ converges weakly to a zero-mean Gaussian process in the metric space $\mathbb{R}^{p+1}\times L^{\infty}(\mathcal{G})$. The limiting covariance matrix of $\sqrt{N}(\hat{\theta}-\theta_0)$ attains the semi-parametric efficiency bound.
\end{thm}

We prove the two theorems in the Appendix. For any $(v, g_{1})\in\mathcal{V}\times\mathcal{G}$, the asymptotic covariance matrix for 
\begin{eqnarray*}
\sqrt{N}v^{\top}\left(\hat{\theta}-\theta_{0}\right)+\sqrt{N}\int_{\cal X} \vec{g}_{1}(x) d\left\{\hat{F}(x)-F_0(x)\right\}
\end{eqnarray*}
can be estimated by $(v^\top, \vec{g_{1}}^{\top}) I_N^{-1}(v^{\top}, \vec{g_{1}}^{\top})^{\top}$, where $NI_N$ is the negative Hessian matrix of the log-likelihood function $l(\theta,{\bf p})$ with respect to $(\theta,{\bf p})$ and $\vec{g_{1}}=(g_{1}\left(x_{1}\right),\ldots, g_{1}\left(x_{N}\right))$. By taking $\vec{g}_{1}=0$, the asymptotic covariance matrix of $\sqrt{N}(\hat{\theta}-\theta_0)$ can be estimated by the upper left $(p+1)\times(p+1)$ matrix of $I_{N}^{-1}$.

Note that in the two theorems we assume that the sample size of the labeled data and the unlabeled data are comparable in the limiting sense. As we have mentioned in Section 1, in many machine learning applications, the size of labeled data is small or moderate while the size of unlabeled data is much larger. When $n/N\to0$ as $n\to\infty$, it can be shown that the convergence rate of $\hat{\theta}$ is at the speed of $\sqrt{n}$ and $\sqrt{n}(\hat{\theta}-\theta_0)$ converges in distribution to a normal distribution with mean zero. The corresponding covariance matrix of the limiting distribution equals to the one obtained from the case-control sample with a known covariate distribution $F$. That is, if $n/N\to0$, we can make inference about $\theta_0$ based on the semi-supervised data as if the marginal distribution of covariates were known to us.

\section{Numerical results}

\subsection{Simulation studies}

We conduct some simulation studies to show the finite sample performance of the proposed method. We consider a 2-dimensional covariates vector $X=(X_1,X_2)$, where $X_1$ are $X_2$ are independently generated from standard normal distribution. The response $Y$ is generated from the logistic model (\ref{M1}) with the parameters $\alpha$, $\beta_1$, and $\beta_2$. The labeled data are drawn by the case-control sampling from the case population and the control population, respectively. The unlabeled data are drawn from the marginal distribution of $X$. We use $\mathsf{p}$ to represent the population percentage of the cases, i.e., $\mathsf{p}=\mathsf{P}(Y=1)$, and use $\mathsf{q}$ to represent the proportion of the cases in the labeled data, i.e., $\mathsf{q}=n_1/n$. 

In the first sets of the study, we concentrate on performance of the estimation procedure. We try different values of the regression parameters which lead to different values of $\mathsf{p}$. We also consider different choices for $n$, $\mathsf{q}$, and $N$. For the generated labeled data and unlabeled data, we use the proposed approach to obtain the MLEs of the regression parameters and the estimated standard errors. At the same time, we use the labeled data only to estimate the parameters, that is, to fit the model (\ref{M1}) with a single case-control sample. One thousand replications are carried out. For each regression parameter, we record the average bias and empirical standard errors of the MLEs, the average of the estimated standard errors, and the empirical coverage percentage of the 95\% Wald confidence interval. We also report the average bias of the estimator of $\mathsf{P}(Y=1)$, which is described in Section 3.2. In Table 1 we list the cases with different regression parameters and the values of $\mathsf{p}$, $n$, and $\mathsf{q}$. The simulation results are listed in Table 2.  

\begin{table}[p]
	\centering
	\caption{Parameter values for data generating and sampling in estimation}
	\label{setting B}
	\bigskip
	\begin{tabular}{cccccccccc}
		\toprule
		Case &$\alpha$ & $\beta_1$ & $\beta_2$ & $n_0$ & $n_1$ & $\mathsf p$ & $\mathsf q$ \\ 
		\midrule
		$A_1$ & $-6$ & $-2$ & 2 & 400 & 80 & 0.037 & 0.167 \\ 
		$A_2$ & $-5$ & $-2$ & 2 & 400 & 80 & 0.067  & 0.167 \\
		$A_3$ & $-4$ & 2 & 1 & 400 & 80 & 0.081  & 0.167 \\ 
		$A_4$ & $-4$ & 2 & 2 & 400 & 80 & 0.116 & 0.167 \\ \\
		$B_1$ & $-6$ & $-2$ & 2 & 300 & 150 & 0.037 & 0.333 \\ 
		$B_2$ & $-5$ & $-2$ & 2 & 300 & 150 & 0.067 & 0.333 \\
		$B_3$ & $-4$ & 2 & 1 & 300 & 150 & 0.081 & 0.333 \\ 
		$B_4$ & $-4$ & 2 & 2 & 300 & 150 & 0.116 & 0.333 \\ \\
		$C_1$ & $-6$ & $-2$ & 2 & 200 & 200 & 0.037 & 0.5 \\ 
		$C_2$ & $-5$ & $-2$ & 2 & 200 & 200 & 0.067 & 0.5 \\ 
		$C_3$ & $-4$ & 2 & 1 & 200 & 200 & 0.081 & 0.5 \\ 
		$C_4$ & $-4$ & 2 & 2 & 200 & 200 & 0.116 & 0.5 \\ 
		\bottomrule
	\end{tabular}
\end{table}

\begin{table}[p]
	\centering
	\caption{Summary results for estimation and inference}
	\label{results B}
	\bigskip
	\scalebox{0.65}{
		\begin{tabular}{ccccccccccccccccc}
			\toprule
			\multirow{2}*{Case} & \multirow{2}*{Type}& \multirow{2}*{N}& \multicolumn{3}{c}{Bias} & \multicolumn{3}{c}{SE} & \multicolumn{3}{c}{ESE} & \multicolumn{3}{c}{CP}&\multirow{2}*{P Bias} \\
			\cmidrule(lr){4-6} \cmidrule(lr) {7-9} \cmidrule(lr) {10-12} \cmidrule(lr) {13-15}
			&  &   & $\alpha$ & $\beta_1$ & $\beta_2$ & $\alpha$ & $\beta_1$  & $\beta_2$ & $\alpha$ & $\beta_1$&$\beta_2$ &$\alpha$ & $\beta_1$&$\beta_2$ &  \\ 
			\midrule
			$A_1$ & single & 480 & 1.553 & $-0.048$ & 0.050 & 0.470 & 0.282 & 0.280 & 0.457 & 0.263 & 0.264 & 0.132 & 0.944 & 0.955 & 0.130  \\  
			& 1 & 960 & $-0.227$  & $-0.071$  & 0.061  & 1.108  & 0.265  & 0.269  & 1.983  & 0.260  & 0.259  & 0.967  & 0.957  & 0.953  & 0.003  \\  
			& 4 & 2400 & $-0.137$  & $-0.059$  & 0.053  & 0.538  & 0.258  & 0.258  & 0.564  & 0.254  & 0.252  & 0.960  & 0.962  & 0.956  & 0.001  \\  
			& 10 & 5280 & $-0.095$  & $-0.021$  & 0.038  & 0.491  & 0.254  & 0.258  & 0.490  & 0.247  & 0.247  & 0.956  & 0.947  & 0.951  & 0.000  \\ \\
			$A_2$ & single & 480 & 0.957 & $-0.036$ & 0.031 & 0.395 & 0.269 & 0.267 & 0.406 & 0.260 & 0.259 & 0.352 & 0.952 & 0.952 & 0.100 \\  
			& 1 & 960 & $-0.100$  & $-0.041$  & 0.056  & 0.557  & 0.266  & 0.269  & 0.548  & 0.258  & 0.259  & 0.951  & 0.951  & 0.946  & 0.003  \\  
			& 4 & 2400 & $-0.093$  & $-0.034$  & 0.039  & 0.464  & 0.254  & 0.255  & 0.464  & 0.254  & 0.255  & 0.968  & 0.951  & 0.947  & 0.001 \\  
			& 10 & 5280 & $-0.078$  & $-0.034$  & 0.044  & 0.446  & 0.260  & 0.259  & 0.446  & 0.254  & 0.254  & 0.951  & 0.948  & 0.955  & 0.002  \\ \\ 
			$A_3$ & single & 480 & 0.779 & 0.026 & 0.021 & 0.262 & 0.232 & 0.178 & 0.298 & 0.235 & 0.183 & 0.261 & 0.959 & 0.963 & 0.086 \\  
			& 1 & 960 & $-0.180$  & 0.038  & 0.028  & 0.847  & 0.247  & 0.185  & 1.440  & 0.235  & 0.181  & 0.954  & 0.943  & 0.951  & 0.000 \\  
			& 4 & 2400 & $-0.078$  & 0.036  & 0.010  & 0.421  & 0.245  & 0.183  & 0.413  & 0.234  & 0.178  & 0.948  & 0.953  & 0.949  & 0.000  \\  
			& 10 & 5280 & $-0.083$  & 0.035  & 0.020  & 0.402  & 0.232  & 0.179  & 0.390  & 0.233  & 0.178  & 0.949  & 0.959  & 0.948  & 0.000  \\  \\
			$A_4$ & single & 480 & 0.362 & 0.039 & 0.034 & 0.340 & 0.258 & 0.262 & 0.358 & 0.258 & 0.258 & 0.774 & 0.953 & 0.954 & 0.051 \\  
			& 1 & 960 & $-0.130$  & 0.045  & 0.056  & 0.471  & 0.273  & 0.270  & 0.466  & 0.258  & 0.259  & 0.949  & 0.945  & 0.954  & $-0.002$ \\  
			& 4 & 2400 & $-0.089$  & 0.042  & 0.048  & 0.421  & 0.262  & 0.259  & 0.417  & 0.258  & 0.258  & 0.952  & 0.957  & 0.950  & $-0.001$  \\  
			& 10 & 5280 & $-0.068$  & 0.043  & 0.056  & 0.410  & 0.255  & 0.255  & 0.406  & 0.258  & 0.259  & 0.954  & 0.957  & 0.957  & 0.001  \\  \\
			$B_1$ & single & 450 & 2.489 & -0.047 & 0.049 & 0.361 & 0.238 & 0.239 & 0.366 & 0.233 & 0.233 & 0.000 & 0.946 & 0.953 & 0.296 \\  
			& 1 & 900 & $-0.336$  & $-0.043$  & 0.041  & 1.385  & 0.230  & 0.222  & 2.469  & 0.219  & 0.219  & 0.971  & 0.940  & 0.959  & 0.000 \\  
			& 4 & 2250 & $-0.127$  & $-0.024$  & 0.021  & 0.565  & 0.206  & 0.209  & 0.646  & 0.205  & 0.205  & 0.964  & 0.950  & 0.954  & 0.000  \\  
			& 10 & 4950 & $-0.095$  & $-0.016$  & 0.021  & 0.371  & 0.202  & 0.211  & 0.381  & 0.197  & 0.198  & 0.965  & 0.958  & 0.935  & $-0.001$ \\  \\
			$B_2$ & single & 450 & 1.864 & -0.047 & 0.041 & 0.311 & 0.229 & 0.228 & 0.327 & 0.230 & 0.230 & 0.005 & 0.957 & 0.952 & 0.267 \\  
			& 1 & 900 & $-0.154$  & $-0.035$  & 0.035  & 0.646  & 0.224  & 0.226  & 0.884  & 0.220  & 0.220  & 0.962  & 0.953  & 0.946  & 0.000  \\  
			& 4 & 2250 & $-0.081$  & $-0.036$  & 0.039  & 0.382  & 0.213  & 0.218  & 0.365  & 0.213  & 0.213  & 0.950  & 0.960  & 0.954  & 0.001 \\  
			& 10 & 4950 & $-0.086$  & $-0.026$  & 0.031  & 0.341  & 0.206  & 0.210  & 0.338  & 0.208  & 0.208  & 0.955  & 0.963  & 0.966  & 0.000 \\  \\
			$B_3$ & single & 450 & 1.708 & 0.029 & 0.018 & 0.217 & 0.210 & 0.159 & 0.234 & 0.206 & 0.159 & 0.000 & 0.941 & 0.951 & 0.253 \\  
			& 1 & 900 & $-0.109$  & 0.035  & 0.013  & 0.476  & 0.208  & 0.153  & 0.490  & 0.202  & 0.151  & 0.968  & 0.945  & 0.945  & 0.000 \\  
			& 4 & 2250 & $-0.061$  & 0.026  & 0.019  & 0.350  & 0.197  & 0.146  & 0.343  & 0.197  & 0.145  & 0.956  & 0.955  & 0.950  & 0.000 \\  
			& 10 & 4950 & $-0.060$  & 0.025  & 0.013  & 0.321  & 0.198  & 0.146  & 0.312  & 0.195  & 0.142  & 0.941  & 0.959  & 0.947  & 0.000 \\  \\
			$B_4$ & single & 450 & 1.304 & 0.039 & 0.033 & 0.274 & 0.237 & 0.229 & 0.283 & 0.227 & 0.226 & 0.021 & 0.946 & 0.943 & 0.217\\  
			& 1 & 900 & $-0.091$  & 0.041  & 0.033  & 0.396  & 0.229  & 0.229  & 0.390  & 0.223  & 0.222  & 0.948  & 0.951  & 0.950  & $-0.001$ \\  
			& 4 & 2250 & $-0.047$  & 0.025  & 0.030  & 0.314  & 0.215  & 0.223  & 0.322  & 0.217  & 0.218  & 0.949  & 0.957  & 0.955  & 0.000  \\  
			& 10 & 4950 & $-0.080$  & 0.033  & 0.029  & 0.318  & 0.219  & 0.221  & 0.308  & 0.216  & 0.216  & 0.946  & 0.951  & 0.944  & $-0.002$ \\  \\
			$C_1$ & single & 400 & 3.190 & $-0.048$ & 0.046 & 0.346 & 0.255 & 0.247 & 0.344 & 0.238 & 0.238 & 0.000 & 0.950 & 0.948 & 0.463 \\  
			& 1 & 800 & $-0.511$  & $-0.051$  & 0.045  & 1.798  & 0.229  & 0.230  & 2.301  & 0.215  & 0.215  & 0.970  & 0.946  & 0.936  & $-0.001$  \\  
			& 4 & 2000 & $-0.166$  & $-0.018$  & 0.025  & 0.492  & 0.193  & 0.185  & 0.476  & 0.193  & 0.193  & 0.962  & 0.951  & 0.961  & $-0.002$ \\  
			& 10 & 4400 & $-0.121$  & $-0.021$  & 0.014  & 0.384  & 0.184  & 0.176  & 0.394  & 0.182  & 0.182  & 0.940  & 0.955  & 0.959  & $-0.002$ \\  \\
			$C_2$ & single & 400 & 2.577 & $-0.031$ & 0.041 & 0.290 & 0.239 & 0.236 & 0.305 & 0.233 & 0.233 & 0.000 & 0.951 & 0.962 & 0.433\\  
			& 1 & 800 & $-0.178$  & $-0.034$  & 0.030  & 0.682  & 0.220  & 0.233  & 0.878  & 0.216  & 0.216  & 0.968  & 0.951  & 0.938  & $-0.001$ \\  
			& 4 & 2000 & $-0.090$  & $-0.029$  & 0.031  & 0.343  & 0.206  & 0.202  & 0.340  & 0.202  & 0.202  & 0.952  & 0.946  & 0.955  & 0.000 \\  
			& 10 & 4400 & $-0.087$  & $-0.026$  & 0.035  & 0.304  & 0.208  & 0.211  & 0.305  & 0.195  & 0.196  & 0.953  & 0.942  & 0.939  & 0.000 \\  \\
			$C_3$ & single & 400 & 2.406 & 0.039 & 0.023 & 0.196 & 0.216 & 0.174 & 0.219 & 0.210 & 0.163 & 0.000 & 0.956 & 0.944 & 0.419  \\  
			& 1 & 800 & $-0.210$  & 0.032  & 0.007  & 0.978  & 0.213  & 0.148  & 2.263  & 0.199  & 0.146  & 0.972  & 0.938  & 0.954  & $-0.002$  \\  
			& 4 & 2000 & $-0.073$  & 0.026  & 0.024  & 0.360  & 0.196  & 0.135  & 0.347  & 0.191  & 0.136  & 0.962  & 0.949  & 0.948  & 0.000 \\  
			& 10 & 4400 & $-0.070$  & 0.028  & 0.014  & 0.296  & 0.192  & 0.130  & 0.291  & 0.186  & 0.131  & 0.948  & 0.947  & 0.956  & $-0.001$ \\  \\
			$C_4$ & single & 400 & 2.002 & 0.024 & 0.039 & 0.254 & 0.230 & 0.233 & 0.265 & 0.230 & 0.231 & 0.000 & 0.951 & 0.956 & 0.384 \\  
			& 1 & 800 & $-0.089$  & 0.022  & 0.033  & 0.401  & 0.217  & 0.213  & 0.383  & 0.218  & 0.219  & 0.950  & 0.955  & 0.958  & $-0.002$  \\  
			& 4 & 2000 & $-0.085$  & 0.038  & 0.035  & 0.288  & 0.220  & 0.223  & 0.296  & 0.211  & 0.211  & 0.957  & 0.948  & 0.944  & $-0.002$  \\  
			& 10 & 4400 & $-0.052$  & 0.026  & 0.033  & 0.277  & 0.211  & 0.209  & 0.274  & 0.206  & 0.206  & 0.952  & 0.950  & 0.961  & $-0.001$  \\  
			\bottomrule
	\end{tabular}}
	{\footnotesize\begin{tablenotes}
			\item[1]Bias: average bias of the estimates; SE: empirical standard error of the estimates; SEE: average of the estimated standard errors; CP: empirical coverage probabilities of Wald-type confidence intervals with $95\%$ confidence level; P Bias: average bias of the estimator of $\mathsf{P}(Y=1)$.
	\end{tablenotes}}
\end{table}

Note that in Table 2, $N=480$ corresponds to the scenario where a single case-control sample is used to fit the logistic model (``single" in the column ``Type"). The numbers in the column ``Type" stands for the ratio between the size of the unlabeled data and that of the labeled data (for example, in case $A_1$, $N=960$ means that the ratio between size of the unlabeled data and that of the labeled data is 1). It has been mentioned that the single case-control sample cannot identify the intercept parameter $\alpha$. Thus, the estimate of $\alpha$ given by the single case-control sample is obviously biased. Meanwhile, the single case-control sample cannot consistently estimate $\mathsf{P}(Y=1)$, either. When the unlabeled data is introduced, the proposed MLE of $\alpha$ is ultimately unbiasd. Moreover, incorporating the unlabeled data also helps in increasing the estimation accuracy of $\beta$. The standard errors of the proposed MLEs of $\beta_1$ and $\beta_2$ are smaller than those obtained from the single case-control sample. Moreover, the standard errors of the proposed MLEs decreases in almost all the scenarios as $N$ increases, especially when $|\mathsf{p}-\mathsf{q}|$ is relatively larger. The proposed estimator of $\mathsf{P}(Y=1)$ also shows reasonable finite sample performance. 

In the second sets of the study, we focus on the prediction performance.  For the regression parameters, we choose the values of Case A1 to A4 listed in Table 1. For the training data, we set $n_0=400$ and $n_1=80$. The size of the unlabeled data is $4800$. Another labeled data set with size $4800$ is generated to serve as the testing data. We consider four methods for prediction. The first two are two existing semi-supervised learning approaches called self-training proposed by Yarowsky (1995) and co-training proposed by Blum and Mitchell (1998), respectively. The Logistic model is used to be the base learner in the two approaches. The third one is to use the labeled data only to train the logistic model. The fourth one is the proposed approach using the MLE for model training. One thousand replications are carried out. We record the average bias of the model parameter estimates and the average AUC (i.e., area under the curve) of the four methods. Moreover, for each method, we use the fitted probability that maximizes the difference between the training true positive rate and false positive rate to be the cut-off point and obtain the corresponding accuracy score (i.e., the percentage of the corrected prediction), recall, precision, and F1 score. Finally, we also record the average absolute bias of the estimation of $\mathsf{P}(Y=1)$ on the testing data. The results are summarized in Table 3.

\begin{table}[p]
	\centering
	\caption{Summary results for prediction}
	\label{tab 3}
	\bigskip
	\scalebox{0.76}{
		\begin{tabular}{ccccccccccc}
			\toprule
			\multirow{2}*{Case} & \multirow{2}*{Method} & \multicolumn{3}{c}{Bias} & \multirow{2}*{AUC} &	\multirow{2}*{accuracy} &\multirow{2}*{recall}
			&\multirow{2}*{precision} & \multirow{2}*{F1 score}	& \multirow{2}*{P Bias} \\
			\cmidrule(lr){3-5} 
			&  &   $\alpha$ & $\beta_1$ &$\beta_2$ & & & & & & \\ 
			\midrule
			
			$A_1$ & self-training & -5.2324  & -2.8577  & 2.8598  & 0.9461  & 0.8658  & 0.8737  & 0.2085  & 0.3339  & 0.0201  \\ 
			& co-training  & 3.7452  & 0.6760  & -1.2364  & 0.9382  & 0.8562  & 0.8729  & 0.1961  & 0.3179  & 0.2891  \\ 
			& case-control & 1.5553  & -0.0508  & 0.0542  & 0.9461  & 0.8653  & 0.8742  & 0.2081  & 0.3333  & 0.0600  \\ 
			& proposed & -0.1100  & -0.0407  & 0.0447  & 0.9462  & 0.8657  & 0.8742  & 0.2085  & 0.3339  & 0.0079  \\ \\
			$A_2$ & self-training & -5.8382  & -3.1816  & 3.1887  & 0.9377  & 0.8539  & 0.8622  & 0.3088  & 0.4510  & 0.0103  \\ 
			& co-training  & 2.8309  & 0.6959  & -1.2123  & 0.9298  & 0.8467  & 0.8574  & 0.2971  & 0.4376  & 0.2628  \\ 
			& case-control & 0.9470  & -0.0423  & 0.0449  & 0.9378  & 0.8535  & 0.8627  & 0.3082  & 0.4504  & 0.0500  \\ 
			& proposed & -0.0711  & -0.0377  & 0.0403  & 0.9378  & 0.8540  & 0.8621  & 0.3091  & 0.4513  & 0.0106  \\ \\
			$A_3$ & self-training & -4.7757  & 3.1405  & 1.5812  & 0.9033  & 0.8159  & 0.8169  & 0.2901  & 0.4239  & 0.0117  \\ 
			& co-training  & 1.7238  & -0.5740  & -0.3010  & 0.8883  & 0.8171  & 0.7866  & 0.2882  & 0.4172  & 0.2477  \\ 
			& case-control & 0.7730  & 0.0411  & 0.0211  & 0.9033  & 0.8151  & 0.8178  & 0.2895  & 0.4233  & 0.0514  \\ 
			& proposed & -0.0680  & 0.0385  & 0.0203  & 0.9034  & 0.8154  & 0.8177  & 0.2897  & 0.4235  & 0.0137  \\ \\
			$A_4$ & self-training & -6.3554  & 3.6116  & 3.6215  & 0.9295  & 0.8415  & 0.8529  & 0.4203  & 0.5588  & 0.0153  \\ 
			& co-training  & 1.9428  & -0.7585  & -1.1693  & 0.9211  & 0.8350  & 0.8443  & 0.4088  & 0.5465  & 0.2177  \\ 
			& case-control & 0.3498  & 0.0509  & 0.0529  & 0.9295  & 0.8422  & 0.8519  & 0.4215  & 0.5597  & 0.0264  \\ 
			& proposed & -0.0818  & 0.0503  & 0.0515  & 0.9295  & 0.8419  & 0.8525  & 0.4209  & 0.5594  & 0.0128  \\ 
			\bottomrule
	\end{tabular}}
	{\footnotesize\begin{tablenotes}
			\item[]Bias: average bias of the estimates; P Bias: average absolute bias of the estimation of $\mathsf{P}(Y=1)$ on the testing data.
	\end{tablenotes}}		
\end{table}

Note that the self-training and co-training approaches are designed for the prospective sampling for the labeled data. For the case-control sampling, the two semi-supervised learning approaches give out similar performances on the AUC, accuracy score, recall, precision, and F1 score compared with the proposed method. However, the two approaches show poor performance on the estimation of the regression parameters in the Logistic model, while the proposed MLE gives out consistent estimates. Consequently, the proposed method has the best prediction accuracy on the marginal percentage $\mathsf{P}(Y=1)$. This is because our proposed method accounts for the biasd sampling scheme in the labeled data. From our results, it seems that ignoring the case-control sampling scheme does not obviously affect the overall accuracy of the individual prediction conditioning on its covariates.

\subsection{Pima Indians diabetes data}

We apply the proposed method to the Pima Indians diabetes data, which is originally from the National Institute of Diabetes and Digestive and Kidney disease. The data set consists of a target variable called ``Outcome" standing for the presence or absence of the diabetes (coded as 1 and 0, respectively) and several medical predictors. Here we use the variables ``Glucose", ``Pregnancies", and ``BMI" as the covariates. There are 268 cases (i.e., ``Outcome" coded as 1) and 500 controls (i.e., ``Outcome" coded as 0) in the data. We randomly split the data into the training set (80\% of the full set) and the testing set (20\% of the full set). In the training set, we randomly draw 50 samples from the cases and 50 samples from the controls to form a case-control sample. Moreover, we randomly draw 200 samples from the training set and use the covariates information to form an unlabeled set. The 300 samples are treated as a semi-supervised data set and the proposed approach is used to get the parameter estimates. Then the fitted logistic model is adopted to do prediction on the testing set. For the $i$th data point in the testing set, we calculate the estimated case probability $\phi(x_i;\hat{\theta})$ and use the mean absolute deviation (MAD) $n_t^{-1}\sum_{i=1}^{n_t}|y_i-\phi(x_i;\hat{\theta})|$, where $n_t$ is the size of the testing set, to measure the prediction accuracy. We also calculate the MAD based on the case-control sample only (i.e., the 50 cases and 50 controls) for comparison. We repeat the sampling procedure for 100 times and get the average of the regression parameter estimates, the average of the estimated standard errors, and the MAD. We treat the parameter estimates obtained from the full training data set and the corresponding MAD for the testing data set as the benchmark. The analysis results are summarized in Table 4.

\begin{table}[p]
	\centering
	\caption{Estimation results for Pima India diabetes data}
	\label{real 2}
	\bigskip
	\begin{tabular}{cccccccc}
		\toprule
		& &\multicolumn{4}{c}{Parameter} & \multirow{2}*{MAD} & \multirow{2}*{Case proportion} \\
		\cmidrule{3-6}
		& & $\alpha$ & $\beta_1$ & $\beta_2$ & $\beta_3$&  &  \\
		\midrule
		Case-control & AEST  & $-0.1925$  & 1.1380  & 0.4084  & 0.7109  & 0.3381  &  \\ 
		& AESE & 0.2513  & 0.3020  & 0.2549  & 0.3063  &  &  \\ \\
		Proposed MLE & AEST & $-0.8971$  & 1.1320  & 0.3942  & 0.7157  & 0.3101  & 0.3403 \\ 
		& AESE & 0.5548  & 0.2995  & 0.2491  & 0.3032  &  &   \\ \\
		Full training set & EST & $-0.7811$  & 1.0518  & 0.3963  & 0.6522  & 0.3126  & \\
		& ESE & 0.0107 & 0.0137 & 0.0098 & 0.0143  &  &   \\
		\bottomrule
	\end{tabular}
	{\footnotesize\begin{tablenotes}
			\item[1]EST.A: average of estimated parameter values; ESE.A: average of estimated standard errors; EST: estimated parameter value; ESE: estimated standard error.
	\end{tablenotes}}
\end{table}

It is quite obvious to see that the average value of the proposed MLE of $\alpha$ is much closer to the MLE based on the full training set than that obtained from the case-control sample. The introduction of the unlabeled data helps in identifying the intercept parameter. Meanwhile, compared with the results based on the single case-control sample, the estimation efficiency of $\beta$ and the prediction accuracy also increase when the unlabeled data are incorporated.

\section{Concluding remarks}

We consider the semi-supervised inference with binary response data under the case-control sampling. Assuming the data following logistic regression model, we find that the unlabeled data helps to identify the intercept parameter which cannot be identified by single case-control data. The likelihood function of the observed data is derived and the MLE of the model parameters are obtained by an iterative algorithm. The MLE is shown to reach the semiparametric efficient bound asymptotically. Numerical studies show that the unlabeled data not only helps in identifying the intercept parameter but also increases the estimation efficiency of the slope parameter. Compared with some existing prediction methods for semi-supervised binary data, the proposed approach give out competitive prediction performances. When the labeled data is case-control data, the proposed approach gives out the most accurate prediction on the marginal case percentage.

In real applications, the models are easily misspecified. For the binary response data, when the logistic model is misspecified, the model parameter becomes a certain population risk minimizer defined through the log-likelihood function of the logistic model. Under this circumstance, the case-control sampling cannot provide a consistent estimator of this risk minimizer. The limit of the MLE may depend on the ratio between the number of cases and that of controls. Although there exist unlabeled data, it seems that the proposed approach does not provide consistent estimator directly for misspecified logistic model. Either it is not clear how the model misspecifiation would affect the performance of prediction. This would be an interesting direction for further investigation in the future. 

\appendix

\section{Appendix}

Here we give out the proof of the two theorems in Section 3.4.

\noindent{\it Proof of Theorem 1}: \ For simplicity of notation, we just consider $p=1$ in this proof and it can be easily extended to a more general space. By definition $\phi(x, \theta)\in(0,1)$ for all $x$. According to Theorem 1 in Vardi (1985), maximizing $l(\theta,\textbf{p})$ yields unique solution for $F(x)$ when $\theta$ is fixed. Denote the jump size $p_i$ by $F\{x_i\}$ and $\hat{p}_i$ by $\hat{F}\{x_i\}$, $i=1,\ldots,N$. Using the method of Lagrange multipliers and some calculus, we can get $$\hat{F}\{x_i\} = \left[\frac{n_1}{\tilde{c}(\hat{\theta}, \hat{F})}\phi(x_i,\hat{\theta})+\frac{n_0}{1-\tilde{c}(\hat{\theta},\hat{F})}\left(1 -\phi(x_i, \hat{\theta})\right)+N-n_1-n_0\right]^{-1},$$ where 
$\tilde{c}(\hat{\theta},\hat{F}) \equiv \tilde{c}(\hat{\theta},\hat{\bf p}) = \sum_{i=1}^N \phi(x_i,\hat{\theta})\hat{F}\{x_i\}$.
Since $\hat{F}$ is uniformly bounded and monotone, for any sub-sequence of $\{\hat{F}\}$, there is a further sub-sequence that converges to some monotone function $F^\star$ point-wise. Without loss of generality, assume that $\hat{\theta}$ converges to $\theta^\star$ for the same sub-sequence.

Denote $$F_{1}(x;\theta, F) = \int_{-\infty}^x \frac{\phi(u, \theta)}{\tilde{c}(\theta,F)}dF(u), \ F_{2}(x; \theta, F) = \int_{-\infty}^x \frac{1-\phi(u, \theta)}{\{1-\tilde c(\theta,F)\}}dF(u)$$ and 
$F_{3}(x; \theta,F )=F_3(x;F) = F(x) $, where $\theta \in \mathcal{B}$ and $F$ is a distribution function.  Let $$\bar{G}_N(x;\theta,F) = \frac{n_1}{N}F_{1}(x; \theta,F) + \frac{n_0}{N}F_{2}(x; \theta, F) + \frac{N-n_1-n_0}{N}F_3(x;F).$$ Then 
\begin{equation}\label{G}
\bar{G}_N(x; \theta,F) = \lambda_{N1}\int_{-\infty}^x \frac{\phi(u,\theta)}{\tilde c(\theta,F)}dF(u)+\lambda_{N0}\int_{-\infty}^x \frac{1- \phi(u, \theta)}{1 - \tilde c(\theta,F)}dF(u) + (1-\lambda_{N1}-\lambda_{N0}) \int_{-\infty}^x dF(u),
\end{equation}
where $\lambda_{N1}=n_1/N$ and  $\lambda_{N0}=n_0/N$. 
Obviously, $\bar{G}_N(x; \theta,F) \to \bar{G}(x; \theta,F) $ as $N\to \infty$, where  
\begin{equation*}
    \bar{G}(x; \theta,F) = \rho_{1}\int_{-\infty}^x \frac{\phi(u,\theta)}{\tilde c(\theta,F)}dF(u)+\rho_{0}\int_{-\infty}^x \frac{1- \phi(u, \theta)}{1 - \tilde c(\theta,F)}dF(u) + (1-\rho_{1}-\rho_{0}) \int_{-\infty}^x dF(u),
    \end{equation*}
$\rho_1 = \lim\limits_{N\to \infty}\lambda_{N1}$ and $\rho_0 = \lim\limits_{N\to \infty}\lambda_{N0}$.
It follows immediately from (\ref{G}) and $\phi(x, \theta) \in (0,1)$ that 
\begin{equation}
\label{F}
F_0(x) = \int_{-\infty}^x \left[\lambda_{N1}\frac{\phi(u, \theta_0)}{\tilde c(\theta_0,F_0)} 
+\lambda_{N0}\frac{1- \phi(u, \theta_0)}{1 - \tilde c(\theta_0,F_0)}+ (1 -\lambda_{N0}-\lambda_{N1})\right]^{-1} d\bar{G}_N(u;\theta_0,F_0).
\end{equation}
Denote 
\begin{equation*}
\mathbb{F}_{N_i}(x) = \left\{
\begin{array}{llr}
n_1^{-1}\sum_{i=1}^{n_1}I\{x_{i}\leqslant x\} &i=1,&\\ \\
n_0^{-1}\sum_{i=n_1 + 1}^{n_0+n_1} I\{x_{i}\leqslant x\}&i= 2,&\\ \\
(N-n_1-n_0)^{-1}\sum_{i=n_0+n_1+1}^{N}I\{x_{i}\leqslant x\}&i = 3. &  
\end{array}
\right.
\end{equation*}
Then we can write 
$$\mathbb{F}_{N}(x) \equiv N^{-1}\sum_{i=1}^{N}I\{x_{i}\leqslant x\}=\lambda_{N1} \mathbb{F}_{N 1}(x)+\lambda_{N0} \mathbb{F}_{N 2}(x)+(1-\lambda_{N1} -\lambda_{N0})\mathbb{F}_{N_3}(x)$$ as the empirical distribution.
Write $\mathbb{X}_N = \sqrt N (\mathbb{F}_N - \lambda_{N1} F_{1}-\lambda_{N0}F_{2}-(1 - \lambda_{N1}-\lambda_{N0})F_{3}) = \sqrt N (\mathbb{F}_N - \bar{G}_N) $. When $\theta = \theta_0$, $F =F_0$ and $\mathcal{F}$ is a Donsker class, 
$\left\{\mathbb{X}_{N}(f): f \in \mathcal{F}\right\}$ converges in distribution in $l^{\infty}(\mathcal{F})$ to the mean zero Gaussian process$\{\mathbb{X}(f): f \in \mathcal{F}\}$.

Then, we consider a class
$$
\begin{aligned}
\mathcal{A}=\bigg\{ &\frac{I_{(-\infty, x]}(X)}{\frac{1}{N}\left[\frac{n_{1}}{\tilde{c}(\theta,F)}\phi\left(X, \theta\right)+\frac{n_{0}}{1-\tilde{c}(\theta,F)}\left\{1-\phi\left(X, \theta\right)\right\} + N - n_1 - n_0\right]}: \theta  \in \mathcal{B},x \in R , \\ 
 & F \text{ is a distribution function}, \tilde{c}(\theta,F)=\sum_{i=1}^{N} \phi\left(x_{i}, \theta\right) 
 F\left\{x_{i}\right\}\bigg\}.
\end{aligned}$$
Similar to Zeng and Yin (2006), It can be shown that $\mathcal{A}$ is a Donsker class. Combining the fact that 
$\mathbb{X}_{N}$ converges to a mean zero Gaussian process and the bounded convergence theorem,
We conclude that uniformly in $x$, $\hat{F}(x)\to F^\star(x)$, where 
$$F^\star(x) = E\left\{\frac{I(X\leq x)}{\rho_1\frac{\phi(X,\theta^\star)}{\tilde{c}(\theta^\star,F^\star)} + \rho_0\frac{\{1 -\phi(X, \theta^\star)\}}{\{1 -\tilde{c}(\theta^\star,F^\star)\}} + 1 -\rho_0 -\rho_1}\right\} $$
and $\mathrm{E}(\cdot)$ refers to the expectation taken with respect to $\bar{G}(x;\theta_0,F_0)$.

We next construct another function $$\tilde{F}\{x_{i}\} = \left[\frac{n_1}{{\tilde c}(\theta_0,F_0)}\phi(x_i,\theta_0) + \frac{n_0}{1 -{\tilde c}(\theta_0,F_0)}\{1 -\phi(x_i, \theta_0)\} + N-n_1-n_0\right]^{-1}.$$
By (\ref{F}) and the fact that $\mathbb{X}_N$ converges to a mean zero Gaussian process, we conclude that $\lim\limits_{N\to\infty}\tilde{F}(x)  = F_0(x)$ uniformly in $x$,  where 
$$\tilde{F}(x) = \frac{1}{N}\sum_{i=1}^{N}\frac{I\{x_{i}\leqslant x\}}{\left[\frac{\lambda_{N1}}{{\tilde c}(\theta_0,F_0)}\phi(x_i,\theta_0) + \frac{\lambda_{N0}}{1 -{\tilde c}(\theta_0,F_0)}\{1 -\phi(x_i,\theta_0)\} + (1 - \lambda_{N1} - \lambda_{N0})\right]}.$$
Note that $\hat{F}(x)$ is absolutely continuous with respect to $\tilde{F}(x)$ with  $\hat{F}(x)$ can be written as 
$$\hat{F}(x) = \int_{-\infty}^x \frac{\frac{n_1}{{\tilde c}(\theta_0,F_0)}\phi(u,\theta_0) + \frac{n_2}{1 -{\tilde c}(\theta_0,F_0)}\{1 -\phi(u,\theta_0)\} + N - n_1 - n_0}{\frac{n_1}{\tilde c(\hat{\theta},\hat{F})}\phi(u,\hat{\theta})+\frac{n_0}{1 - \tilde c(\hat{\theta},\hat{F})}\{1 -\phi(u,\hat{\theta}) \}+N-n_1-n_0} d \tilde F(u).$$
Then     $$F^\star(x) = \int_{-\infty}^x \frac{\frac{\rho_1}{{\tilde c}(\theta_0,F_0)}\phi(u,\theta_0) + \frac{\rho_2}{1 -{\tilde c}(\theta_0,F_0)}\{1 -\phi(u,\theta_0)\} + 1 - \rho_1 - \rho_0}{\frac{\rho_1}{\tilde{c}(\theta^\star,F^\star)}\phi(u,\theta^\star)+\frac{\rho_0}{1 - \tilde{c}(\theta^\star,F^\star)}\{1 -\phi(u,\theta^\star) \}+1 -\rho_1-\rho_0} d F_0(u).$$
This implies that $F^\star(x)$ is absolutely continuous with respect to $F_0(x)$ and 
$$\lim\limits_{n \rightarrow \infty} \frac{\hat{F}\{x\}}{\tilde{F}\{x\}}=\frac{f^{*}(x)}{f^\star_0(x)},$$
where $f^\star$ and $f^\star_0$ are the density functions of $F^\star$ and $F_0$ respectively.

Finally, We show that $\theta^\star = \theta_0$ and $F^\star = F_0.$ Using the fact that $l_n(\hat{\theta}, \hat{F}) \ge l_n(\theta_0, \tilde{F})$, we have that
$$\begin{aligned}
&\frac{1}{N}\sum_{i=1}^{n_1}\log{\frac{\phi(x_i,\hat{\theta})}{\phi(x_i,\theta_0)}} + \frac{1}{N}\sum_{i=n_1+1}^{n_0+n_1}\log{\frac{1 -\phi(x_i,\hat{\theta})}{1 -\phi(x_i,\theta_0)}} -\frac{n_1}{N}\log\frac{c(\hat{\theta},\hat{F})}{\tilde{c}(\theta_0, \tilde{F})} -\\
& \frac{n_0}{N}\log\frac{1 -c(\hat{\theta},\hat{F})}{1-\tilde{c}(\theta_0, \tilde{F})} 
+ \frac{1}{N}\sum_{i=1}^{N}\log\frac{\hat{F}\{x_{ik}\}}{\tilde{F}\{x_{ik}\}} \ge 0.
\end{aligned}
$$
Taking limits on both sides and defining $$A_1(\theta,F) =\log \frac{\phi(x,\theta)}{\tilde{c}(\theta,F)}f(x), \ A_2(\theta, F) = \log \frac{1 -\phi(x,\theta)}{1 -\tilde{c}(\theta, F)}f(x),$$ and $A_3(F) =\log f(x)$, where $f(x)$ is the density function of $F(x)$, we obtain that
$$\rho_1 E_1\left[\frac{A_1(\theta^\star, F^\star)}{A_1(\theta_0, F_0)}\right] + \rho_0 E_0\left[\frac{A_2(\theta^\star, F^\star)}{A_2(\theta_0, F_0)}\right] + (1-\rho_1-\rho_0)E_X\left[\frac{A_3(F^\star)}{A_3(F_0)}\right] \ge 0, $$ 
where $\rho_1 = \lim\limits_{N\to \infty}\lambda_{N1}$ , $\rho_0 = \lim\limits_{N\to \infty}\lambda_{N0}$, $\mathrm{E}_{1}(\cdot)$ refers to the expectation taken with respect to $f_{1}(x)$,
$\mathrm{E}_{0}(\cdot)$ refers to the expectation taken with respect to $f_{0}(x)$ and 
$\mathrm{E}_{X}(\cdot)$ refers to the expectation taken with respect to the marginal density of $X$.
Thus the left-hand side is the weighted sum of three negative Kullback-Leibler distance. Therefore, the identifiability condition {\it C2} requires that $\theta^\star= \theta_0$ and $F^\star = F_0$ with probability one.
Furthermore, the continuity of $F_0$ ensures that the convergence is uniform in $\cal X$.\qed

\medskip

\noindent{\it Proof of Theorem 2}: \ The proof is based on the argument on MLE of Van der Vaart (2000). We first introduce the set 
$$\mathcal{H} = \{(v, g): v\in\mathbb{R}^{d+1}, \vert\vert v \vert\vert \leqslant  1,g \in BV[d], |g| \leqslant 1\}.$$
Let $\boldsymbol{P}_{n}$ be the empirical measure of $n$ i.i.d observations, that is, $\boldsymbol{P}_{n}\{g(X)\}=n^{-1}\sum_{i=1}^{n} g\left(X_{i}\right)$. Denote $\boldsymbol{P}\{g(X)\}=\mathsf{E}[g(X)],$ where $g(\cdot)$ is an arbitrary measurable function. Separate samples of fixed size are drawn from three sub-populations in our setting. Similar to Breslow et al. (2000), we can modify this setting so that it involves a simple random sample of size $N$ from a biased sampling model. Specifically, consider the semiparametric biased sampling model $\mathbf{P} = \left\{P_{\theta,F}; \theta \in \Theta,F\in \mathbf{F}\right\}$, defiened on $\mathcal{Z} =\{0,1,2\}\times \mathcal{X}$ with density
given by 
\begin{equation*}
p(z,\theta, F) = p(i, x, \theta, F)  = \left\{
\begin{array}{llr}
\frac{n_0}{N}\frac{\{1-\phi(x,\theta)\}}{1-c(\theta,F)}f(x) &i=0,&\\ \\
\frac{n_1}{N}\frac{\phi(x,\theta)}{c(\theta,F)}f(x)&i= 1,&\\ \\
\frac{N-n}{N}f(x)&i = 2. &  
\end{array}
\right.
\end{equation*}

Furthermore, we abbreviate $l(\theta,F)$ as $\log p(z,\theta,F)$. Let $l_{\theta}(\theta, F)$ be the derivative of $l(\theta, F)$ with respect to $\theta$ along the direction of $\theta+\varepsilon v,$ and let $l_{F}(\theta, F)\left[g \right]$ be the derivative of $l(\theta, F)$ with respect to $F$ along the direction of $F_{\varepsilon}=F+\varepsilon \int g du$. Moreover, we denote the derivative of $l_{\theta}(\theta, F)$ with respect to $\theta$ as $l_{\theta \theta}(\theta, F)$. Similarly, we denote the derivative of $l_{\theta}(\theta, F)$ with respect to $F$ as $l_{\theta F}(\theta, F)\left[ g \right]$ and denote the derivative of $l_{F}(\theta, F)\left[g\right]$ with respect to $F$ along the path $F+\varepsilon\left(\hat{F}-F\right)$ as $ l_{F F}(\theta, F)\left[ g, \hat{F}-F\right]$.

By the definition of $(\hat{\theta},\hat{F})$, we have that for any  $\left(v, g\right) \in \mathcal{H}$, $$P_{N}\left\{l_{\theta}\left(\hat{\theta}, \hat{F}\right) v^{\top}+l_{F}\left(\hat{\theta}, \hat{F}\right)\left[g \right]\right\}=0.$$
Note that $P\left\{l_{\theta}\left(\theta_{0}, F_{0}\right) v^{\top}+l_{F}\left(\theta_0, F_0\right)\left[g\right]\right\}=0$. Thus, we obtain that
\begin{eqnarray}\label{PN}
& &\sqrt{N}\left(\boldsymbol{P}_{N}-\boldsymbol{P}\right)\left\{l_{\theta}\left(\hat{\theta}, \hat{F}\right) v^{\top}+l_{F}\left(\hat{\theta}, \hat{F}\right)\left[g\right]\right\}\nonumber \\
&=&-\sqrt{N} \boldsymbol{P}\left\{l_{\theta}\left(\hat{\theta}, \hat{F}\right) v^{\top}+l_{F}\left(\hat{\theta}, \hat{F}\right)\left[g\right]\right\}+\sqrt{N} \boldsymbol{P}\left\{l_{\theta}\left(\theta_{0}, F_{0}\right) v^{\top}+l_{F}\left(\theta_{0}, F_{0}\right)\left[g\right]\right\}.
\end{eqnarray} 

Define $$\mathcal{A}_{1}=\left\{l_{\theta}(\theta, F) v^{\top}+l_{F}(\theta, F)\left[g\right]:\|v\| \leq 1, \int g d x=0, F+\int \varepsilon g d u \geq 0\right\}.$$
According to the explicit expression of $l_{\theta}(\theta, F),l_{F}(\theta, F) $ and the preservation of the Donsker class under algebraic operations, we have that $\mathcal{A}_1$ is a Donsker class for any positive $\epsilon$. Using the Donsker theorem, we have that 
\begin{equation}\label{leftside}
\sqrt{N}\left(\boldsymbol{P}_{N}-\boldsymbol{P}\right)\left\{l_{\theta}\left(\theta_{0}, F_{0}\right) v^{\top}+l_{F}\left(\theta_{0}, F_{0}\right)\left[g\right]\right\}+o_{p}(1).
\end{equation}
By using Talyor expansion, the right hand side of (\ref{PN}) can be formulated as 
\begin{eqnarray}
& &l_{\theta}\left(\theta_{0}, F_{0}\right) v^{\top}+l_{F}\left(\theta_{0}, F_{0}\right)\left[g\right]-l_{\theta}\left(\hat{\theta}, \hat{F}\right) v^{\top}-l_{F}\left(\hat{\theta}, \hat{F}\right)\left[g\right]\nonumber\\
&=&\left(\hat{\theta}-\theta_{0}\right) l_{\theta \theta}\left(\theta_{0}, F_{0}\right) v^{\top}+l_{F F}\left(\theta_{0}, F_{0}\right)\left[g, \hat{F}-F_{0}\right]+\left(\hat{\theta}-\theta_{0}\right) l_{F \theta}\left(\theta_{0}, F_{0}\right)\left[g\right] v^{\top}+v^{\top} l_{F \theta}\left[\hat{F}-F_{0}\right]\nonumber\\
& &+o\left(\left\|\hat{\theta}-\theta_{0}\right\|+\left\vert\hat{F}-F_{0}\right\vert\right).
\end{eqnarray}
Then, the right side of (\ref{PN}) equals to
\begin{equation*}
\sqrt{N}\left\{\left(\hat{\theta}-\theta_{0}\right) \Omega_{\theta}\left(v,g\right)+\int \Omega_{F}\left(v, g\right) d\left(\hat{F}-F_{0}\right)\right\}+o\left\{\sqrt{N}\left(\left\|\hat{\theta}-\theta_{0}\right\|+\left\vert\hat{F}-F_{0}\right\vert\right)\right\},
\end{equation*}
where $\Omega_{\theta}\left(v, g\right)$ is the operator
$$\Omega_{\theta}\left(v, g\right)=P\left\{l_{\theta \theta}\left(\theta_{0}, F_{0}\right) v^{\top}+l_{F \theta}\left(\theta_{0}, F_{0}\right)\left[g\right]\right\}$$
and
$$\Omega_{F}\left(v, g\right)=P\left\{v^{\top} l_{\theta F}\left(\theta_{0}, F_{0}\right)+l_{F F}\left(\theta_{0}, F_{0}\right)\left[g, \hat{F}-F_{0}\right]\right\}.$$

According to Theorem 4.7 in Rudin (1973), the information operator is invertible if and only if it is one-to-one. To show that $\left(\Omega_{\theta}, \Omega_{F}\right)$ is injective, suppose that $\Omega_{\theta}\left(v,g\right)=0$ and $\Omega_{F}\left(v, g\right)=0$. Then
$$\Omega_{\theta}\left(v, g\right) v^{\top}+\int \Omega_{F}\left(v, g\right) g d F_0=0.$$
The derivation of $\Omega's$ tell us that
\begin{equation*}
\Omega_{\theta}\left(v, g\right) v^{\top}+\int \Omega_{F}\left(v, g\right) g d F_0=-E\left\{l_{\theta}\left(\theta_{0}, F_{0}\right)^{T}v+l_{F}\left(\theta_{0}, F_{0}\right)\left[g\right]\right\}^{2}.
\end{equation*}
Hence, $l_{\theta}\left(\theta_{0}, F_{0}\right) v^{\top}+l_{F}\left(\theta_{0}, F_{0}\right)\left[g \right]=0$ almost surely. By condition {\it C2}, $\left(v^{\top}, g\right)=0$ and $\operatorname{ker}\left(\Omega_{\theta}, \Omega_{F}\right)=0,$ then the operator is surjective. As a result, $\left(\Omega_{\theta}, \Omega_{F}\right)$ is one-to-one and thus invertible. 
Now with  $\left(\tilde\Omega_{\theta}\left(v, g\right), \tilde\Omega_{F}\left(v, g\right)\right)=\left(\Omega_{\theta}, \Omega_{F}\right)^{-1}\left(v, g\right) .$ Consequently, expression (\ref{PN}) can be written as
\begin{eqnarray}\label{convergence}
& &\sqrt{N}\left\{\left(\hat{\theta}-\theta_{0}\right) v^{\top}+\int g d\left(\hat{F}-F_{0}\right)\right\}+o\left\{\sqrt{N}\left(\left\|\hat{\theta}-\theta_{0}\right\|+\left\vert\hat{F}-F_{0}\right\vert\right)\right\}\nonumber\\
&=& \sqrt{N}\left(\boldsymbol{P}_{N}-\boldsymbol{P}\right)\left\{l_{\theta}\left(\theta_{0}, F_{0}\right) \tilde\Omega_{\theta}\left(v, g\right)^{\top}+l_{F}\left(\theta_{0}, F_{0}\right) \int \tilde\Omega_{F}\left(v, g\right) d F_{0}\right\}.
\end{eqnarray}
It follows immediately from (\ref{leftside}) that the right hand side of (\ref{convergence}) converges in distribution to a zero-mean Gaussian process. By the Slutsky's theorem, we now only need to show that 
\begin{equation}
\label{order}
\sqrt{N}\left(\left\|\hat{\theta}-\theta_{0}\right\|+\left\vert\hat{F}-F_{0}\right\vert\right) = O_p(1)
\end{equation}
By definition,
$$\left|\left|\hat{\theta}-\theta_{0}\right|\right|+\left|\hat{F}-F_{0}\right| = \sup\limits_{(v, g) \in \mathcal{H} }\left|v^{T}\left(\hat{\theta}-\theta_{0}\right)+\int_X g d\left(\hat{F}-F_{0}\right)\right|
$$.
Thus, (\ref{convergence}) gives that
$$\sqrt N\left|\left|\hat{\theta}-\theta_{0}\right|\right|+\sqrt N \left|\hat{F}-F_{0}\right| = O_p(1) +  o\left(\sqrt N \left\| \hat{\theta}-\theta_0\right\| + \sqrt N\left|\hat{F}-F_0\right|\right).$$
Therefore, (\ref{order}) holds immediately. Now we have that
\begin{eqnarray}\label{convergence1}
& &\sqrt{N}\left\{\left(\hat{\theta}-\theta_{0}\right) v^{\top}+\int g d\left(\hat{F}-F_{0}\right)\right\}\nonumber\\
&=& \sqrt{N}\left(\boldsymbol{P}_{N}-\boldsymbol{P}\right)\left\{l_{\theta}\left(\theta_{0}, F_{0}\right) \tilde\Omega_{\theta}\left(v, g\right)^{\top}+l_{F}\left(\theta_{0}, F_{0}\right) \int \tilde\Omega_{F}\left(v, g\right) d F_{0}\right\} + o_p(1).
\end{eqnarray}
Then $\sqrt{N}\left(\hat{\theta}-\theta_{0}\right) v^{\top}+\int g d\left(\hat{F}_{N}-F_{0}\right)$ converges weakly to a Gaussian process. Finally, when $g=0$ in (\ref{convergence}), the estimator $\hat{\theta}$ is asymptotically linear with influence function $l_{\theta}\left(\theta_{0}, F_{0}\right) \tilde\Omega_{\theta}^{\top}+l_{F}\left(\theta_{0}, F_{0}\right) \int \tilde\Omega_{F} d F_{0}$, which lies in the linear space spanned by the
score functions 
$$\Big\{l_{\theta}\left(\theta_{0}, F_{0}\right) \tilde\Omega_{\theta}\left(v, 0\right)^{\top}+l_{F}\left(\theta_{0}, F_{0}\right) \int\tilde\Omega_{F}\left(v,0\right)
d F_{0}:\left(\tilde\Omega_{\theta}\left(v, g\right), \tilde\Omega_{F}\left(v, g\right)\right) \in \mathcal{H}\Big\}.$$ It follows from Proposition
1 in Bickel {\it et al}. (1993) that $\hat{\theta}$ is semiparametric efficient. This completes the proof of Theorem 2.\qed

\bigskip

\noindent {\bf Acknowledgment}

Yuanyuan Lin's research is partially supported by the Hong Kong Research Grants Council (Grant No.14306219 and 14306620), Direct Grants for Research, The Chinese University of Hong Kong. Kani Chen's research is supported by Hong Kong General Research Fund 16212117. Wen Yu's research is supported by the National Natural Science Foundation of China Grants (12071088). 

\medskip

\bibliographystyle{plain}

\end{document}